\RequirePackage[svgnames]{xcolor}

\documentclass[11pt,letterpaper]{mystyle}

\usepackage[comma,authoryear,compress]{natbib}
\usepackage{algorithm}
\usepackage{algorithmicx}
\usepackage{algpseudocode}
\usepackage{microtype}
\usepackage{graphicx}
\usepackage{booktabs}
\usepackage{multirow}
\usepackage{array}
\usepackage{float}
\usepackage{bigstrut}
\usepackage{amsmath}
\usepackage{amssymb}
\usepackage{mathtools}
\usepackage{amsthm}
\usepackage{mathrsfs}
\usepackage{nicefrac}
\usepackage{dsfont}
\usepackage{enumitem}
\usepackage{subcaption}
\usepackage{cleveref}
\usepackage[font=small,labelfont=bf]{caption}
\usepackage{color}
\usepackage{adjustbox}
\usepackage{rotating}
\usepackage{makecell}
\usepackage{bxcoloremoji}

\newcommand{\x}{\mathbf{x}}

\definecolor{blanchedalmond}{rgb}{1.0, 0.92, 0.8}
\definecolor{carmine}{rgb}{0.59, 0.0, 0.09}
\definecolor{lightblue}{rgb}{0.22,0.45,0.70}%

\renewcommand{\mathbf}{\boldsymbol}

\makeatletter
\def\Ddots{\mathinner{\mkern1mu\raise\p@
\vbox{\kern7\p@\hbox{.}}\mkern2mu
\raise4\p@\hbox{.}\mkern2mu\raise7\p@\hbox{.}\mkern1mu}}
\makeatother

\definecolor{amaranth}{rgb}{0.9, 0.17, 0.31}
\definecolor{antiquebrass}{rgb}{0.8, 0.58, 0.46}
\definecolor{antiquefuchsia}{rgb}{0.57, 0.36, 0.51}
\definecolor{chromeyellow}{rgb}{0.31, 0.47, 0.26}

\definecolor{lightblue}{rgb}{0.22,0.45,0.70}
\definecolor{Gray}{gray}{0.95}
\definecolor{Cornsilk}{rgb}{1.0, 0.97, 0.86}

\title{EgoCS-400K: An Egocentric Gameplay Dataset for World Models}

\runningtitle{EgoCS-400K}

\author{
Rongjin Guo$^{\ast}$, Dong Liang$^{\ast}$, Yuhao Liu$^{\dagger}$, Fang Liu, Tianyu Huang,  Gerhard P. Hancke, and Rynson W. H. Lau\\
City University of Hong Kong\\
{\small $^{\ast}$Equal contribution. $^{\dagger}$Corresponding author.}
}

\begin{document}

\begin{abstract}
The shift from video generation to interactive world modeling places new demands on data: beyond captioned videos, world models require temporally aligned video-action-language trajectories grounded in the actions, camera motion, states, and events that drive future scene changes. However, such data is difficult to obtain at scale. Web video datasets offer broad visual coverage but lack executable actions and reliable states; robotic datasets provide action and state supervision but are costly and limited in scene diversity; and existing simulators often lack large-scale human-driven interaction trajectories. In this paper, we introduce \textbf{EgoCS-400K}, a large-scale replay-grounded egocentric Counter-Strike dataset for world models, built from public professional CS and CS2 match demos that preserve human gameplay trajectories and enable parsing, replaying, rendering, and temporal alignment. We extract player states, view directions, movements, keyboard/button inputs, view-angle changes, weapon usage, game events, and round-level context, and render clean first-person videos from the same trajectories. EgoCS-400K contains over \textbf{400,000} first-person videos and \textbf{10,000} hours of gameplay from more than \textbf{1,000} matches and \textbf{40,000} rounds, covering \textbf{13} maps and \textbf{10} player viewpoints per round. It supports a range of interactive visual modeling tasks, including action-conditioned future prediction, state- and event-aware scene rollout, replay-grounded captioning, and agent egocentric action understanding. By connecting visual observations with human actions, camera motion, game states, and events at scale, EgoCS-400K serves as a practical bridge between passive web videos, controllable game simulation, and costly real-world embodied data.

\vspace{2mm}


\vspace{5mm}

\textbf{Date}: June 16, 2026

\textbf{Project}: \href{https://EgoCS-400K.github.io}{https://EgoCS-400K.github.io}

\textbf{Contact}: \href{https://yuhaoliu7456.github.io/}{Yuhao LIU, yuhaoliu7456@gmail.com}

\end{abstract}

\sloppy
\maketitle
\vspace{3mm}

\begin{figure}[!t]
    \centering
    \includegraphics[width=\linewidth]{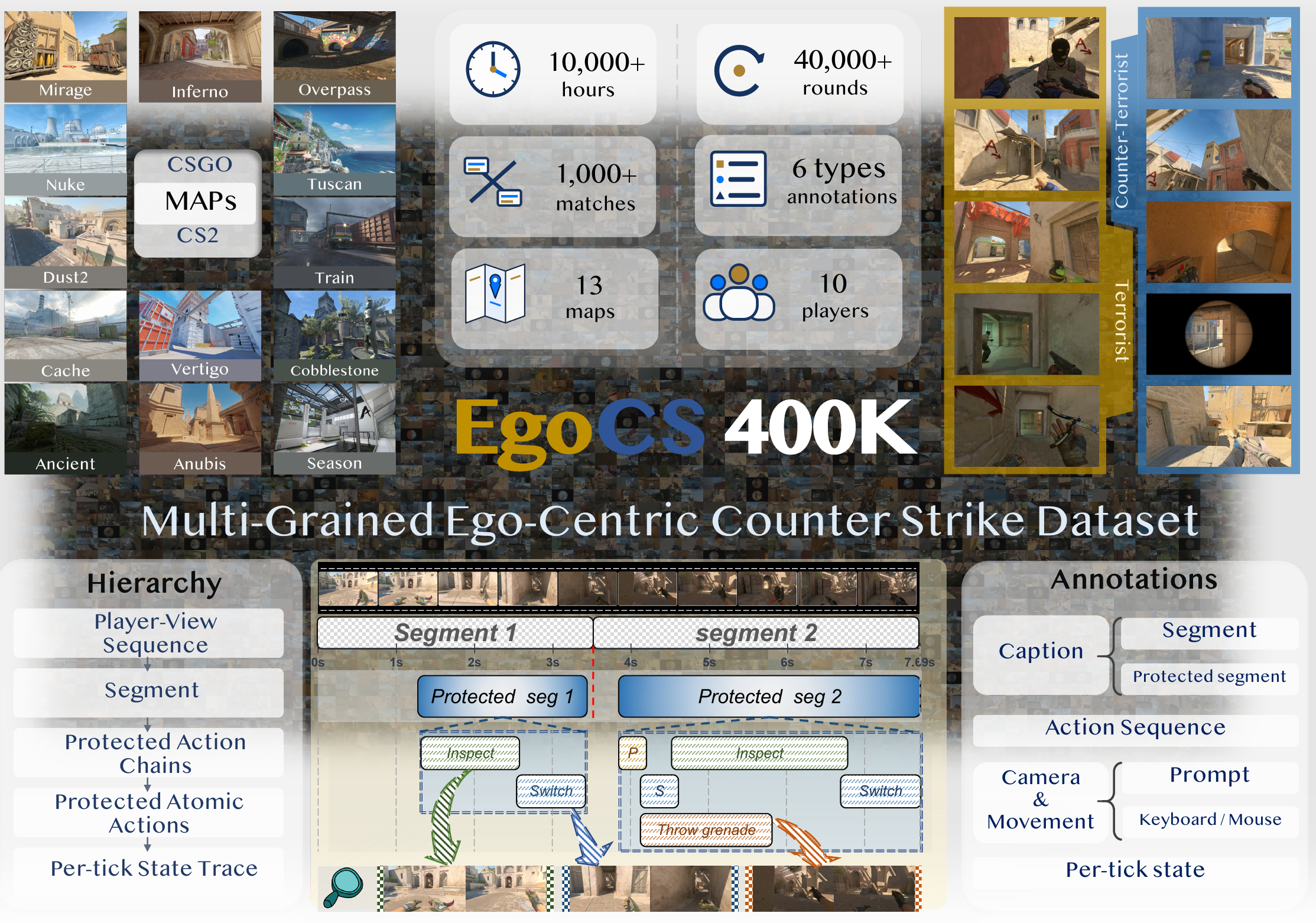}
    \caption{
    \textbf{Overview of EgoCS-400K.}
    EgoCS-400K is a large-scale multi-grained egocentric Counter-Strike dataset built from first-person gameplay videos across CS:GO and CS2 maps, covering 10{,}000+ hours, 40{,}000+ rounds, 1{,}000+ matches, 13 maps, and 10 player viewpoints. 
    The dataset provides a hierarchical organization from player-view sequences to segments, protected action chains, protected atomic actions, and per-tick state traces. 
    For each segment, EgoCS-400K further provides rich replay-grounded annotations, including segment-level and protected-segment captions, action sequences, camera and movement descriptions, prompts, keyboard/mouse signals, and per-tick states. 
    This multi-level design enables fine-grained analysis of embodied actions, temporal dynamics, and player-environment interactions in complex competitive scenarios.
    }
    \label{fig:teaser}
\end{figure}

\section{Introduction}

Recent advances in large-scale video generation models \citep{brooks2024sora,polyak2024moviegen}, interactive world models \citep{bruce2024genie,valevski2024gamengen}, and vision-language-action models \citep{sima2,kim2024openvla,black2024pi0} are shifting visual modeling from producing realistic videos toward understanding how actions change the world. This shift requires models to learn visual dynamics, i.e., how observations evolve over time, and action-conditioned dynamics, where future observations depend on control signals, camera motion, and decisions made by an embodied agent. In egocentric settings, models must further understand first-person perception and how language corresponds to actions, temporal segments, states, and scene changes. This transition fundamentally changes the data needed for training visual models: passive videos and weak captions are no longer sufficient, while video-language-action trajectories become increasingly important.

However, such trajectories are difficult to obtain at scale in the real world. Large web video datasets provide broad visual coverage, but their supervision is mostly passive and their language is only weakly aligned with the actions that cause visual changes \citep{miech2019howto100m,bain2021frozen,schuhmann2022laion5b}. Egocentric video datasets capture first-person human experience across diverse activities, but they usually lack precise control traces, reliable internal states, and temporally grounded event records \citep{grauman2022ego4d,damen2018epickitchens,damen2022rescaling}. Robotic datasets provide action and state supervision, but they are costly to collect and often limited in embodiment, scene diversity, and interaction complexity. Games and simulators offer a practical bridge: they are scalable and repeatable, while exposing actions, states, camera trajectories, and environment events. Prior game-based datasets have demonstrated the value of human demonstrations and time-aligned modalities for embodied AI \citep{guss2019minerl,fan2022minedojo,he2025plaicraft}. Nevertheless, existing game and simulation datasets are not specifically designed for world models.

To build such data, we seek a source that is human-driven, first-person, visually rich, temporally precise, and replay-grounded. Counter-Strike demos provide a natural fit: each demo records a full replay-grounded human gameplay trajectory, including player states, camera motion, actions, game events, and round-level context. Unlike ordinary gameplay videos, demos preserve the underlying trajectory from which visual observations can be replayed, rendered, and temporally aligned. This replay-grounded structure makes Counter-Strike demos an attractive source for egocentric video-language-action data, enabling models to learn how observations evolve under human actions in dynamic environments.

Based on this insight, we introduce \textbf{EgoCS-400K}, a multi-grained egocentric Counter-Strike dataset for world models. EgoCS-400K is built from public professional CS:GO and CS2 match demos, which provide large-scale human gameplay trajectories without requiring new manual data collection. We parse each replay with DemoParser2 to extract player states, view directions, movements, weapon usage, utility events, combat events, and round-level context. The same replay trajectories are rendered into clean first-person videos, producing visual observations temporally aligned with the parsed signals. We further organize the data into hierarchical temporal units, including player-centric sequences, video segments, protected action chains, protected atomic actions, and per-tick state traces. For each segment, replay-derived facts constrain the action, state, camera, and event annotations, while visual frames supplement appearance, environment, and scene-level details for captions and prompts.

As shown in Fig.~\ref{fig:teaser}, EgoCS-400K contains more than \textbf{10,000 hours} of first-person video from over \textbf{1,000} matches and \textbf{40,000} rounds, yielding over \textbf{400,000} first-person videos by rendering \textbf{10} player viewpoints per round, covering \textbf{13} maps. This design supports a range of world-modeling-related tasks, including action-conditioned future prediction, state- and event-aware scene rollout, controllable egocentric video simulation and agent egocentric action understanding. Although Counter-Strike is a simulated environment, it captures transferable egocentric priors such as navigation, visual search, camera control, partial observability, multi-agent interaction, and action-conditioned scene evolution. These properties make EgoCS-400K a practical intermediate testbed between passive web-scale videos and expensive real-world embodied data.

Our contributions are as follows:
\begin{itemize}[leftmargin=*]
\item We introduce \textbf{EgoCS-400K}, a large-scale replay-grounded egocentric Counter-Strike dataset built from public CS:GO and CS2 demos, providing clean first-person videos aligned with human actions, camera motion, player states, and game events.
\item We develop a multi-grained annotation pipeline that organizes gameplay into player-view sequences, video segments, protected action chains, protected atomic actions, and per-tick state traces, with captions and prompts constrained by replay-derived facts.
\item We provide a practical testbed for interactive visual modeling, supporting action-conditioned future prediction, state- and event-aware scene rollout, controllable egocentric video simulation and agent egocentric action understanding.
\end{itemize}

\section{Related Work}

\textbf{Interactive world models.} Recent interactive world models increasingly treat video generation as controllable world simulation rather than passive future prediction. A central requirement is action-conditioned control, where future frames are generated from user inputs, latent actions, keyboard/mouse commands, or high-level instructions \citep{bruce2024genie,oasis2024,valevski2024diffusion,alonso2024diffusion,yu2025gamefactory,guo2025mineworld,tang2025hunyuangamecraft,tang2025hunyuangamecraft2,mao2025yume,mao2025yume15}. Another requirement is real-time or streaming interaction, so that the model can continuously respond to changing controls instead of producing isolated clips \citep{feng2024matrix,zhang2025matrixgame,wang2026matrixgame3}. Recent work also emphasizes long-horizon consistency, memory, and geometric grounding, which are necessary for persistent worlds, repeated scene visits, and camera-aware generation \citep{xiao2025worldmem,hong2025relic,shang2025longscape,nam2026worldcam,worldscape2026}. In parallel, explorable 3D world generation aims to construct persistent and navigable virtual environments from images or text, further highlighting the need for world-consistent visual dynamics and camera-aware supervision \citep{huang2025voyager,team2025hunyuanworld}. Together, these directions show that interactive video world models require dense supervision beyond video-text pairs: precise controls, camera motion, state transitions, and event-level outcomes are needed for scalable training and reliable evaluation. EgoCS-400K addresses this data gap by converting replay files into clean first-person videos with synchronized controls, camera motion, internal game state, weapon and utility events, and language supervision, providing replay-auditable data for multishot interactive world modeling.

\noindent \textbf{Egocentric video datasets.} Ego4D\citep{grauman2022ego4d} established large-scale first-person video as a foundation for long-horizon perception, social interaction, hand-object reasoning, and episodic memory . EPIC-KITCHENS and its later expansion focus on daily kitchen activities with rich action labels and narrations \citep{damen2018epickitchens,damen2022rescaling}. These datasets are visually and behaviorally diverse, but their action labels are not the low-level controls that generate future observations. EgoCS-400K is complementary: it is less semantically open-ended than real-world egocentric video, but every video segment is tied to replay-derived controls, camera motion, internal game state, and discrete game events.

\noindent \textbf{Video-language and action recognition datasets.} Kinetics \citep{kay2017kinetics}, ActivityNet \citep{heilbron2015activitynet}, UCF101 \citep{soomro2012ucf101}, Something-Something \citep{goyal2017something}, and YouCook2 \citep{zhou2018youcook2} helped standardize video understanding tasks around action classification, temporal localization, or procedural description. HowTo100M \citep{miech2019howto100m} and WebVid \citep{bain2021frozen} scale video-text learning through web supervision, while image-language datasets such as COCO \citep{lin2014coco}, Visual Genome \citep{krishna2017visualgenome}, and LAION-5B \citep{schuhmann2022laion5b} illustrate how scale and weak captions can drive representation learning. These resources are valuable for passive perception, retrieval, and captioning, but they generally do not expose the action/state trace that caused the video. EgoCS-400K instead treats language as one layer of a structured replay-grounded record.

\noindent\textbf{Game and embodied-agent datasets.}
Game environments have long served as controllable testbeds for embodied learning and generalist virtual-world agents \citep{sima2}.
MineRL provides Minecraft demonstrations for imitation and reinforcement learning \citep{guss2019minerl}; MineDojo combines Minecraft gameplay with Internet-scale knowledge and open-ended tasks \citep{fan2022minedojo}; PLAICraft records large-scale Minecraft interactions with time-aligned video, audio, speech, mouse, and keyboard modalities \citep{he2025plaicraft}. Robotics datasets such as DROID \citep{khazatsky2024droid} and Open X-Embodiment \citep{padalkar2023openx} similarly emphasize large-scale trajectories with action supervision. EgoCS-400K follows the same principle of paired observation-action data, but differs in two respects: it reconstructs supervision from replay files rather than instrumented live logging, and it targets high-fidelity first-person video suitable for generative world modeling.

\noindent \textbf{Multimodal data engines and structured supervision.}
Large datasets increasingly rely on automated or semi-automated data engines. Segment Anything pairs model-assisted annotation with a massive segmentation corpus \citep{kirillov2023sam}; autonomous-driving datasets such as nuScenes \citep{caesar2020nuscenes}, Waymo Open Dataset \citep{sun2020waymo}, and BDD100K \citep{yu2020bdd100k} combine synchronized sensors, temporal structure, and task-specific labels. EgoCS-400K uses a replay data engine: the demo file is the source of truth, rendering produces the first-person observation stream, and parsing produces dense annotations. This design gives the dataset a natural audit path from video segment back to replay event.

\section{Method}
\label{sec:method}

\paragraph{Overview.}
EgoCS-400K is designed as an egocentric video-language-action dataset with temporally aligned and source-traceable annotations. Its core design principle is to decouple canonical supervision from visual observation: the Counter-Strike demo timeline provides authoritative timing, player state, actions, camera motion, weapon state, and game-event signals, while the rendered first-person video provides the corresponding visual evidence for scene description and caption generation. This design makes each segment auditable across the original demo timeline, rendered video interval, and derived action-state annotations.

\begin{figure*}[t]
    \centering
    \includegraphics[width=\linewidth]{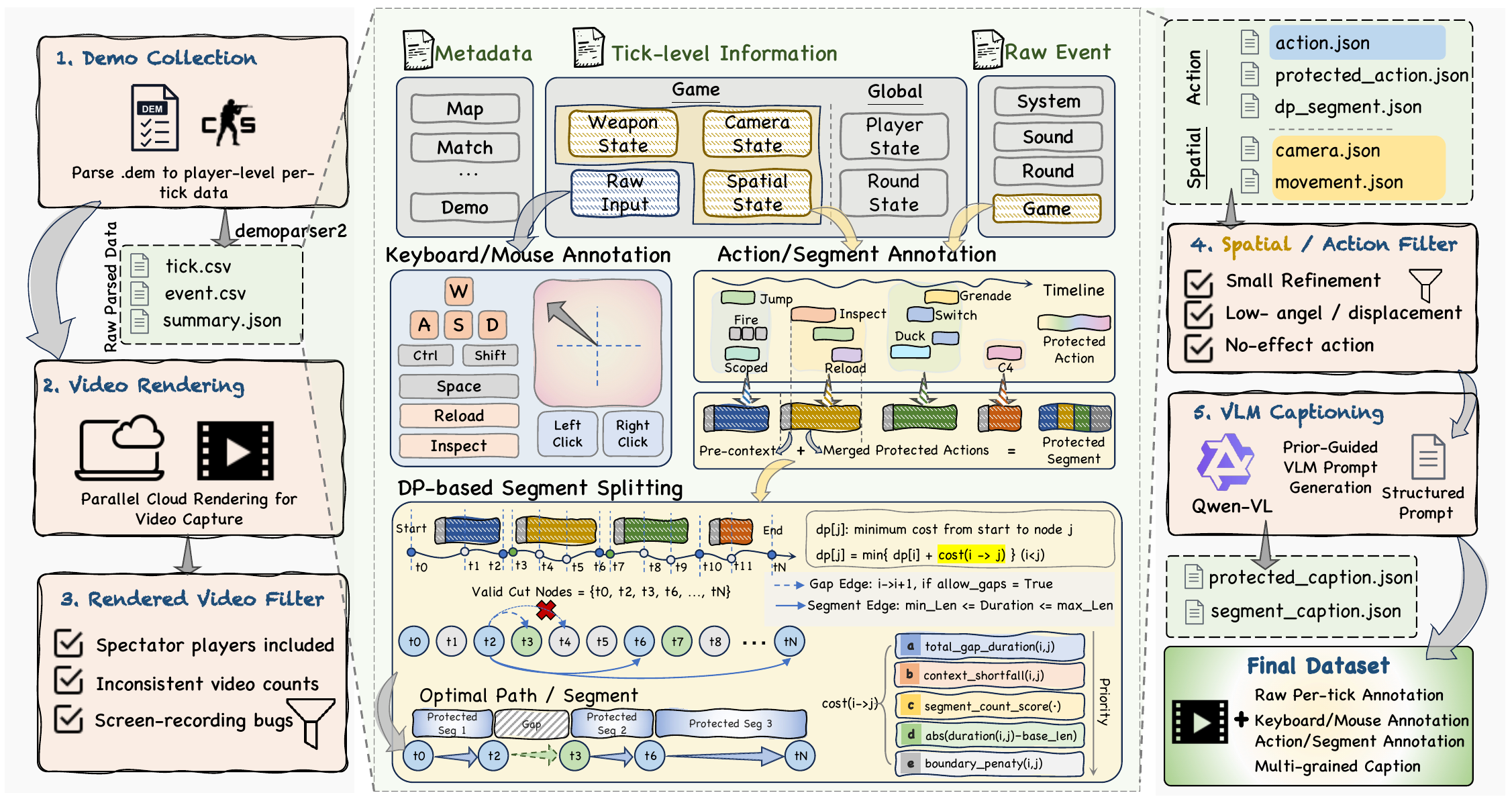}
    \caption{
    \textbf{Construction pipeline of EgoCS-400K.}
    Starting from public Counter-Strike demos, we first collect raw replay files, render synchronized first-person videos, and filter invalid rendered videos.
    We then parse each demo into player-level per-tick data.
    From the parsed action and spatial signals, we construct keyboard/mouse annotations, atomic action sequences, protected action chains, and video segments.
    Finally, we construct prior-guided VLM prompts to generate structured segment-level and protected-chain-level captions.
    }
    \label{fig:pipeline}
\end{figure*}

As shown in Fig.~\ref{fig:pipeline}, the construction process consists of five illustrated steps, which we group into three methodological phases. In the first phase, we collect demos, render first-person videos, and filter invalid recordings. In the second phase, we parse each demo into player-level tick traces and atomic action spans, then organize these spans into action-safe temporal segments. In the third phase, we construct prior-guided VLM inputs by filtering action, movement, and camera priors before generating structured captions. The resulting dataset provides synchronized per-tick state traces, keyboard/mouse signals, action and segment annotations, and multi-grained captions. Table~\ref{tab:schema} summarizes the annotation levels produced by the pipeline and the supervision exposed at each level.

\begin{table}[h]
  \centering
  \caption{Multi-level annotation schema. Each level is derived from the same replay timeline and can be mapped to video time.}
  \label{tab:schema}
  \small
  \begin{tabularx}{\linewidth}{@{}l l X@{}}
    \toprule
    Level & Artifact & Supervision \\
    \midrule
    Tick state & \texttt{ticks.csv} & Controls, view angles, position, velocity, and states \\
    \midrule
    Atomic actions & \texttt{events.csv} & Fire, reload, switch, inspect, scope, grenade, crouch \\
    \midrule
    Action timeline &
    \begin{tabular}[t]{@{}l@{}}
      \texttt{action.json} \\
      \texttt{protected\_action.json}
    \end{tabular}
    & Frame-level actions and protected chains \\
    \midrule
    Training segments & \texttt{dp\_segments.json} & DP-planned clip boundaries and included actions \\
    \midrule
    Captions &
    \begin{tabular}[t]{@{}l@{}}
      \texttt{segment\_caption.json} \\
      \texttt{protected\_caption.json}
    \end{tabular}
    & Structured scene draft and long prompt \\
    \bottomrule
  \end{tabularx}
\end{table}

\subsection{Demo Collection, Rendering, and Filtering}
\label{sec:method_collection}

\textbf{Demo collection.}
We collect public professional CS:GO and CS2 match demos from HLTV. Each demo serves as the authoritative source of temporal information, including round boundaries, player identities, positions, view angles, input states, weapon states, utility trajectories, combat events, and global match events. Starting from a curated match list, our pipeline resolves demo identifiers, retrieves the corresponding files, and maintains a ledger of completed, skipped, and failed entries to support reproducible and resumable data collection.

\textbf{First-person rendering.}
We generate first-person videos from demos through a metadata-guided rendering process. For each demo, we first extract round and player metadata, including the map, player identity, and round tick range. Each rendering task is defined by a tuple of demo, round, player, and tick interval. We render the selected interval from the target player's first-person viewpoint using CS Demo Manager and the Counter-Strike client. Since CS2 demos run at 64 ticks per second, videos are rendered at 32 FPS, allowing each frame to be aligned with a deterministic tick interval. The resulting videos are encoded as H.264 MP4s with CRF 23 and preserved audio, and are stored together with match, round, player, sequence, and tick metadata.

\textbf{Rendered-video filtering.}
Rendered videos are filtered before annotation to remove failures introduced by demo incompatibility, non-standard demo files, or long-running rendering jobs. Specifically, we discard captures that break player-view alignment or corrupt the visual distribution, including spectator-view recordings, failed screen recordings, missing or inconsistent player views, and clips whose duration or metadata does not match the parsed tick interval. The remaining videos are treated as clean first-person observations for downstream parsing, segmentation, and captioning.

\subsection{Parsing and Segmentation}
\label{sec:method_parsing}

\textbf{Player-level parsing.}
For each demo, the parser detects playable rounds from replay events such as \texttt{round\_freeze\_end} and \texttt{round\_end}. Within each round, it exports player-specific artifacts under a consistent match/round/player directory. The primary per-tick artifact is \texttt{ticks.csv}, which records frame, tick, time, player identity, position, pitch/yaw, view-angle deltas, mouse-delta proxies, velocity, movement buttons, fire/right-click/reload/use inputs, ground and duck states, active weapon, ammo, scoped state, reload state, inspect state, and weapon animation state. A \texttt{summary.json} file stores the round id, sequence id, tick range, player identity, weapons seen, event counts, and action counts.

\textbf{Keyboard and mouse reconstruction.}
Keyboard and mouse annotations are derived from synchronized per-tick state rather than manual labeling. Discrete key and button states are decoded from the demo button bitmask, covering movement keys, jump, duck, walk, fire, right-click, reload, and use. For mouse motion, we use view-angle changes as continuous proxies. Given pitch \(\theta_t\) and yaw \(\psi_t\), we compute
\begin{equation}
\Delta \theta_t = \theta_t - \theta_{t-1}, \qquad
\Delta \psi_t = \operatorname{wrap}_{[-180,180)}(\psi_t - \psi_{t-1}),
\label{eq:view_angle_delta}
\end{equation}
and store
\begin{equation}
\texttt{mouse\_dy}_t = \Delta \theta_t, \qquad
\texttt{mouse\_dx}_t = \Delta \psi_t .
\label{eq:mouse_proxy}
\end{equation}
These quantities measure per-tick view-angle displacement rather than raw hardware mouse counts, and are used for visualization, movement/camera prior extraction, and prompt construction.

\textbf{Atomic action extraction.}
The parser converts dense tick traces and game events into temporally bounded atomic action spans stored in \texttt{events.csv}. Each span records its action type, subtype, tick and video-time interval, player identity, relevant weapon or item, source signal, confidence, end reason, and a structured details payload. Rather than relying on manual action labels, we implement rule-based detectors that map synchronized raw signals to action spans. For example, weapon switches are detected from active-weapon changes, reloads from continuous reload-state flags, inspection from weapon animation states, and ducking actions from duck-state flags and duck amount.

These rule-based spans form the basic action units used by the timeline builder, protected-chain construction, segment planning, and prompt generation. Multi-stage actions are represented as temporally linked sub-events when the underlying signals support them. For instance, grenade usage is inferred from the joint evidence of active grenade state, fire or right-click button holds, weapon-fire events, \texttt{grenade\_thrown} events, projectile-flight intervals, and effect events such as detonation or smoke expiration. This produces separate spans for preparation, release, flight, and effect while preserving their relation as one high-level grenade action.

\textbf{Action timeline and protected chains.}
The atomic spans in \texttt{events.csv} are extracted independently from different source signals, so they may be short, dense, and temporally overlapping. To make these spans usable for visualization and video segmentation, we first normalize them into a unified action timeline. Each span is mapped from demo ticks to video frames using the round start tick, tick rate, and video FPS. The resulting timeline records frame-level action intervals and packed display lanes for visualization.

We then derive cut-protected intervals from the action timeline. An action is marked as cut-protected when splitting through it would create a visible or semantic discontinuity, such as interrupting a weapon draw, reload, grenade preparation or flight, or scope transition. In contrast, persistent environmental effects, such as lingering grenade effects, and state-only spans such as sustained scoped or crouched intervals, are not treated as cut-protected player actions. Finally, overlapping or adjacent cut-protected intervals are merged into non-overlapping protected chains. Each protected chain defines the minimal continuous interval that should remain intact during video segmentation.

\textbf{Dynamic-programming segmentation.}
Player-view sequences are often much longer than the temporal units needed for captioning, prompting, and downstream video modeling. We therefore divide each sequence into short training segments while preserving action-chain integrity. Fixed-length slicing is insufficient because it ignores these temporal constraints and can therefore yield semantically incomplete clips. Under these constraints, segmentation becomes a boundary-selection problem: segment boundaries should cover the player-view sequence while never falling inside a protected interval. We represent the video as an ordered set of valid boundary nodes. A node is valid if it is not strictly inside any protected interval. Let \(V=\{t_0,\ldots,t_N\}\) be the sorted valid nodes from the start to the end of the video. This formulation naturally yields a dynamic program. Each candidate segment corresponds to an edge \(i\rightarrow j\) between two valid nodes, and the edge is allowed only when the duration \(t_j-t_i\) falls within the configured segment-length range. When gaps are enabled, we additionally allow gap edges only between adjacent valid nodes. Because edge validity depends only on the two boundary nodes and the protected intervals, and because the total segmentation cost is additive over selected edges, the optimal segmentation has the standard optimal-substructure property. We compute
\begin{equation}
    D[j] = \operatorname*{lexmin}_{i<j,\,(i,j)\in\mathcal{E}}
    \left(D[i] + C(i,j)\right),
\end{equation}
where \(\mathcal{E}\) is the set of allowed segment and gap edges. The edge cost \(C(i,j)\) is a lexicographically ordered cost vector:
\begin{equation}
    C(i,j)=\left(G(i,j),\,P(i,j),\,N(i,j),\,\left|(t_j-t_i)-T\right|,\,B(i,j)\right).
\end{equation}
Here \(G(i,j)\) penalizes uncovered gaps, \(P(i,j)\) denotes the pre-action context shortfall, \(N(i,j)\) encodes the segment-count preference, \(T\) is the target segment duration, and \(B(i,j)\) penalizes less preferred boundaries. We include \(P(i,j)\) because the visual evidence needed to describe an action often begins before the protected interval itself: view direction, approach motion, hand or weapon preparation, and surrounding context make the subsequent action more interpretable for captioning and video modeling. The default configuration uses compact segmentation with a 2.0s minimum length, 4.0s target length, 6.5s maximum length, and 0.5s desired pre-action context. The resulting \texttt{dp\_segments.json} stores segment boundaries, included protected-chain indices, included action ids, unsegmented gaps, settings, and diagnostics such as over-long protected chains.

\subsection{Prior-Guided VLM Captioning}
\label{sec:method_captioning}

Captions are not treated as free-form video summaries in EgoCS-400K. They provide language supervision that connects first-person visual evidence with the action, movement, camera, and state traces used by downstream video generation and VLA-style learning. Generic video captions are insufficient for this setting: they often describe only salient scene appearance or outcomes, remain weakly aligned with the underlying control timeline. Unconstrained VLM captions may also hallucinate unsupported game events or describe persistent contextual effects as player actions. We therefore generate prior-guided captions at two granularities. Segment-level captions describe each DP-selected clip as a complete video unit, emphasizing scene progression, camera motion, movement, visible actions, and final state. Protected-chain-level captions target dense action intervals when such chains exist, using a shorter temporal window to describe fine-grained action continuity.

\subsubsection{Segment-local prior construction.}
For each caption target, we convert the global annotation timeline into a local prompt instance. For segment-level captions, the target window is a DP-selected training segment; for protected-chain-level captions, the target window is a dense protected action interval. In both cases, tick traces, action spans, movement events, camera events, and state summaries are clipped to the target window and re-based to local time. This local representation lets the VLM reason over the visual evidence and the aligned priors within the same temporal frame, without exposing unrelated actions from the surrounding player-view sequence.

\subsubsection{Prior filtering.}
The segment-local structured priors passed to the VLM are organized into three roles. \emph{Action priors} specify player-executed events that should be reflected in the temporal description. \emph{Movement priors} summarize locomotion patterns inferred from keyboard states and player displacement. \emph{Camera priors} summarize view-direction changes inferred from pitch and yaw deltas. Other visual information, such as map geometry, lighting, occlusion, smoke, and fire, is supplied by the clipped video rather than by structured priors.

This filtering step addresses two opposite failure modes. Without structured priors, the VLM often misses short or mechanically subtle actions that are visually brief but important, such as weapon switches, recoil recovery, or small camera turns. However, passing every parsed signal to the VLM is also undesirable. Overly dense priors can create spurious visual grounding, where the model incorrectly associates a minor noisy motion with a salient visual object. For example, if a small leftward view jitter immediately precedes a clear right turn that reveals a doorway, an unfiltered camera prior may cause the model to incorrectly attach the doorway to the leftward motion. We therefore filter priors to retain temporally meaningful action, movement, and camera cues while suppressing low-evidence or visually negligible signals.

\textbf{Action priors.} For action priors, we keep player-executed events that are expected to produce visible temporal changes, such as weapon switching, firing, reloads, inspections, grenade preparation and flight, scope transitions, melee actions, and short posture or airborne transitions. We remove \texttt{noEffect} actions because they correspond to parsed inputs or spans without reliable visible effect. For example, a fire input issued during a weapon-switch or draw animation does not necessarily produce anactual shot, so it should not be described as firing in the caption.

\textbf{Movement and camera priors.}
Movement and camera priors are constructed from the segment-local tick trace rather than from the action timeline. For movement, we first group contiguous W/A/S/D states into temporal runs. For a run \(r=[a,b]\), we compute its planar displacement and mean speed as
\begin{equation}
    d_r = \sqrt{(x_b-x_a)^2 + (y_b-y_a)^2}, \qquad
    \bar{v}_r = \frac{1}{b-a+1}\sum_{t=a}^{b} v^{2D}_t .
\end{equation}
A non-stationary key run is treated as ineffective motion if both \(d_r < \tau_d\) and \(\bar{v}_r < \tau_v\). We further merge unstable short runs and assign the remaining movement prior according to displacement: low-displacement runs are suppressed, moderate-displacement runs are described as small position adjustments, and high-displacement runs are mapped to directional labels such as forward movement, backward movement, or strafing.

For camera motion, we aggregate the per-tick view-angle displacements defined in Eq.~\ref{eq:view_angle_delta}. For a temporal bin \(b\), we compute yaw and pitch statistics as
\begin{equation}
    \Delta^{\psi}_b = \sum_{t\in b}\Delta\psi_t, \qquad
    A^{\psi}_b = \sum_{t\in b}|\Delta\psi_t|,
\end{equation}
\begin{equation}
    \Delta^{\theta}_b = \sum_{t\in b}\Delta\theta_t, \qquad
    A^{\theta}_b = \sum_{t\in b}|\Delta\theta_t|.
\end{equation}
Bins with small absolute displacement are discarded as camera jitter. Contiguous active bins are then merged into candidate view events. For an event \(e\) on axis \(u\in\{\psi,\theta\}\), we compute
\begin{equation}
    \Delta^{u}_e = \sum_{b\in e}\Delta^{u}_b, \qquad
    A^{u}_e = \sum_{b\in e}A^{u}_b, \qquad
    \rho^{u}_e = \frac{|\Delta^{u}_e|}{A^{u}_e}.
\end{equation}
The event is retained as a directional camera prior only when \(A^{u}_e\) and \(|\Delta^{u}_e|\) exceed the angular-motion thresholds and \(\rho^{u}_e\) indicates a consistent direction. Yaw events are expressed as left/right turns, and pitch events as looking down or raising the view. These priors constrain viewpoint continuity without forcing the caption to repeat mechanical input labels.

\subsubsection{Prompt and output schema.}
The final VLM request contains the clipped video, segment-relative action priors, movement priors, camera priors, and a deterministic temporal skeleton. The temporal skeleton orders the required action, movement, and camera facts, ensuring that the generated caption follows the same temporal progression as the structured priors. The prompt instructs the VLM to treat these priors as constraints, while using the video to fill in visual details such as hands, weapon appearance, map geometry, lighting, occlusion, and visible environmental effects. During the inference stage, the VLM is required to return strict JSON with four top-level fields: \texttt{scene\_draft}, \texttt{long\_prompt}, \texttt{confidence}, and \texttt{flags}. The \texttt{scene\_draft} organizes the caption into first-person visual details, environment progression, visible effects, and chronological events. The \texttt{long\_prompt} converts this structured draft into a coherent video-generation caption. This output format keeps the caption temporally grounded while still allowing the model to add visual details that are observable in the video but absent from the structured priors.
\section{Analysis and Limitations}
\label{sec:analysis_limitations}

\subsection{Dataset Overview}
\label{sec:dataset_overview}

EgoCS-400K is a large-scale egocentric video-language-action dataset built from professional CS:GO and CS2 gameplay. Its basic unit is a round-player video, corresponding to the first-person view of one player within one playable round. At release scale, EgoCS-400K contains more than 400K round-player videos, over 10K hours of first-person video, more than 40K rounds, over 1K professional matches, 13 maps, and up to 10 synchronized player viewpoints per round. The average round-player video length is approximately 90 seconds (Table~\ref{tab:dataset_overview}).

Beyond video, EgoCS-400K provides dense temporally aligned annotations for each player-view sequence. These annotations include per-tick player states, keyboard and mouse signals, weapon and movement states, atomic action spans, action timelines, protected chains, training segments, and multi-grained captions. Segment-level captions describe complete clip-level visual progression, while protected-chain-level captions focus on dense action intervals. All annotations share the same tick-based temporal reference, allowing each segment and caption to be traced back to the corresponding video interval, action spans, and player-state trajectory.

\begin{table}[t]
\centering
\small
\caption{Release-scale overview of EgoCS-400K.}
\label{tab:dataset_overview}
\begin{tabular}{lcccccc}
  \toprule
  & Matches & Rounds & Round-player videos & Total video & Avg. length &
  Maps \\
  \midrule
  EgoCS-400K & $>$1,000 & $>$40,000 & $>$400,000 & $>$10,000 h & $
  \approx$90.0 s & 13 \\
  \bottomrule
\end{tabular}
\end{table}

This design makes EgoCS-400K a scalable testbed for action-conditioned modeling and agent-oriented learning. Compared with ordinary video-caption datasets, EgoCS-400K exposes the underlying action, camera, state and event signals that drive visual change. This enables models to study how first-person observations evolve under human actions, including navigation, viewpoint control, visual search, tactical interaction, and rapid action transitions. Although the environment is game-based, the dataset provides a practical bridge toward real-world egocentric and embodied-agent settings where dense temporal supervision is difficult to obtain at scale.

\subsection{Qualitative Samples}
\label{sec:qualitative_samples}

\begin{figure}[h]
  \centering
  \includegraphics[width=\linewidth]{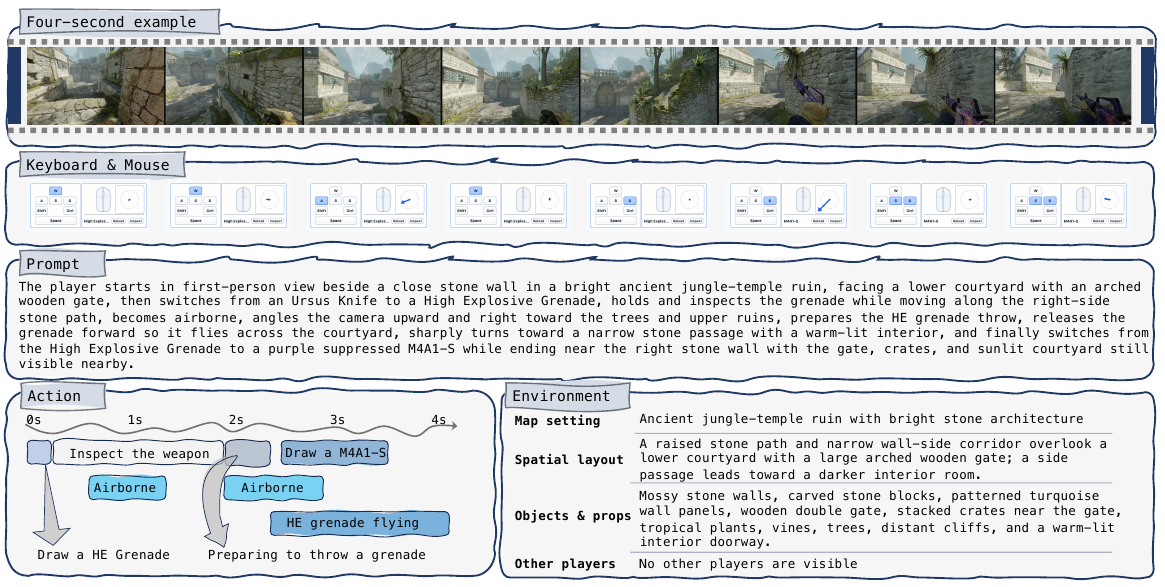}
  \caption{A four-second qualitative example. The visualization aligns first-person frames, input traces, action annotations, and the generated prompt within the same segment.}
  \label{fig:four_second_example}
\end{figure}

Figure~\ref{fig:four_second_example} shows a four-second example with synchronized visual and annotation layers. The visualization includes sampled first-person frames, keyboard and mouse traces, action timeline, and the generated prompt for the same temporal window. This example illustrates how EgoCS-400K connects short visual changes with action structure: weapon switching, inspection, airborne movement, grenade preparation, projectile flight, and the subsequent weapon state are temporally ordered and aligned with the rendered frames. The resulting caption therefore provides a multi-dimensional description of the segment, covering player actions, movement, camera motion, weapon state, and surrounding environment. These dimensions are not presented as separate lists; instead, they are fused into a coherent temporally evolving caption that describes how the first-person scene changes over time.

Figure~\ref{fig:four_second_example} also visualizes the temporal action sequence within the segment, making the order and duration of player actions and states easier to inspect. The action timeline and input traces expose the structured temporal supervision, while the VLM-generated environment description complements it with detailed visual context, including surrounding geometry, lighting, occlusion, and visible scene elements.

\subsection{Limitations}
\label{sec:limitations}

EgoCS-400K is a densely supervised testbed for action-conditioned egocentric video modeling and embodied-agent research. Its behavioral distribution is centered on Counter-Strike gameplay, where the dominant patterns involve tactical navigation, viewpoint control, target engagement, weapon and utility handling, and rapid action transitions. These behaviors are useful for studying first-person dynamic modeling, but they do not cover the full range of everyday activities, fine-grained hand-object manipulation, social interaction, household tasks, or open-ended real-world behavior. The action space is also shaped by Counter-Strike mechanics. Player controls are discrete keyboard and mouse inputs, weapons and utility items follow game-defined state machines, and scene changes obey game physics and map design rather than physical-world dynamics. These properties make large-scale temporal supervision feasible, but they also mean that EgoCS-400K should be viewed as an intermediate testbed rather than a direct model of real-world embodiment. The main domain gaps lie in continuous physical interaction, tactile feedback, deformable or manipulable objects, and non-combat everyday behavior. Finally, the captions are prior-guided VLM annotations rather than manually written gold labels. The structured priors constrain action, movement, and camera facts, and the generated outputs include confidence and flags to support auditing and filtering. Nevertheless, the captions may still contain visual-detail errors or imperfect grounding. EgoCS-400K should therefore be viewed as a scalable intermediate testbed for densely supervised egocentric dynamics and transfer toward broader embodied settings.

\section{Conclusion}

EgoCS-400K uses competitive first-person gameplay as a scalable setting for video-language-action data. By pairing clean rendered videos with dense annotations, the dataset fills an existing gap in large-scale egocentric data by aligning visual observations, language descriptions, actions, states and event structure within a shared timeline. This design supports fine-grained analysis of how first-person scenes evolve under human actions, including embodied navigation, active visual perception, action-conditioned temporal dynamics, and long-horizon human behavior.

Beyond dataset construction, EgoCS-400K provides a practical testbed for action-conditioned generation, and agent-oriented egocentric modeling. While the game-based
setting introduces domain gaps to real-world embodiment, it offers a scalable bridge toward models that learn the temporal coupling between visual observations and the actions that drive them.

\clearpage
\bibliography{main}

@inproceedings{grauman2022ego4d,
  title={Ego4D: Around the World in 3,000 Hours of Egocentric Video},
  author={Grauman, Kristen and Westbury, Andrew and Byrne, Eugene and Chavis, Zachary and Furnari, Antonino and Girdhar, Rohit and Hamburger, Jackson and Jiang, Hao and Liu, Miao and Liu, Xingyu and Martin, Miguel and Nagarajan, Tushar and Radosavovic, Ilija and Ramakrishnan, Santhosh Kumar and others},
  booktitle={Proceedings of the IEEE/CVF Conference on Computer Vision and Pattern Recognition},
  year={2022}
}

@inproceedings{damen2018epickitchens,
  title={Scaling Egocentric Vision: The EPIC-KITCHENS Dataset},
  author={Damen, Dima and Doughty, Hazel and Farinella, Giovanni Maria and Fidler, Sanja and Furnari, Antonino and Kazakos, Evangelos and Moltisanti, Davide and Munro, Jonathan and Perrett, Toby and Price, Will and Wray, Michael},
  booktitle={Proceedings of the European Conference on Computer Vision},
  year={2018}
}

@article{damen2022rescaling,
  title={Rescaling Egocentric Vision: Collection, Pipeline and Challenges for EPIC-KITCHENS-100},
  author={Damen, Dima and Doughty, Hazel and Farinella, Giovanni Maria and Furnari, Antonino and Ma, Jian and Kazakos, Evangelos and Moltisanti, Davide and Munro, Jonathan and Perrett, Toby and Price, Will and Wray, Michael},
  journal={International Journal of Computer Vision},
  year={2022}
}

@inproceedings{kay2017kinetics,
  title={The Kinetics Human Action Video Dataset},
  author={Kay, Will and Carreira, Joao and Simonyan, Karen and Zhang, Brian and Hillier, Chloe and Vijayanarasimhan, Sudheendra and Viola, Fabio and Green, Tim and Back, Trevor and Natsev, Paul and Suleyman, Mustafa and Zisserman, Andrew},
  booktitle={arXiv preprint arXiv:1705.06950},
  year={2017}
}

@inproceedings{heilbron2015activitynet,
  title={ActivityNet: A Large-Scale Video Benchmark for Human Activity Understanding},
  author={Heilbron, Fabian Caba and Escorcia, Victor and Ghanem, Bernard and Niebles, Juan Carlos},
  booktitle={Proceedings of the IEEE Conference on Computer Vision and Pattern Recognition},
  year={2015}
}

@article{soomro2012ucf101,
  title={UCF101: A Dataset of 101 Human Actions Classes From Videos in the Wild},
  author={Soomro, Khurram and Zamir, Amir Roshan and Shah, Mubarak},
  journal={CRCV-TR-12-01},
  year={2012}
}

@misc{sima2,
      title={SIMA 2: A Generalist Embodied Agent for Virtual Worlds}, 
      author={SIMA team},
      year={2025},
      eprint={2512.04797},
      archivePrefix={arXiv},
      url={https://arxiv.org/abs/2512.04797}, 
}

@inproceedings{goyal2017something,
  title={The Something Something Video Database for Learning and Evaluating Visual Common Sense},
  author={Goyal, Raghav and Ebrahimi Kahou, Samira and Michalski, Vincent and Materzynska, Joanna and Westphal, Susanne and Kim, Heuna and Haenel, Valentin and Fruend, Ingo and Yianilos, Peter and Mueller-Freitag, Moritz and others},
  booktitle={Proceedings of the IEEE International Conference on Computer Vision},
  year={2017}
}

@inproceedings{zhou2018youcook2,
  title={Towards Automatic Learning of Procedures From Web Instructional Videos},
  author={Zhou, Luowei and Xu, Chenliang and Corso, Jason J.},
  booktitle={Proceedings of the AAAI Conference on Artificial Intelligence},
  year={2018}
}

@inproceedings{miech2019howto100m,
  title={HowTo100M: Learning a Text-Video Embedding by Watching Hundred Million Narrated Video Clips},
  author={Miech, Antoine and Zhukov, Dimitri and Alayrac, Jean-Baptiste and Tapaswi, Makarand and Laptev, Ivan and Sivic, Josef},
  booktitle={Proceedings of the IEEE/CVF International Conference on Computer Vision},
  year={2019}
}

@inproceedings{bain2021frozen,
  title={Frozen in Time: A Joint Video and Image Encoder for End-to-End Retrieval},
  author={Bain, Max and Nagrani, Arsha and Varol, Guel and Zisserman, Andrew},
  booktitle={Proceedings of the IEEE/CVF International Conference on Computer Vision},
  year={2021}
}

@inproceedings{lin2014coco,
  title={Microsoft COCO: Common Objects in Context},
  author={Lin, Tsung-Yi and Maire, Michael and Belongie, Serge and Hays, James and Perona, Pietro and Ramanan, Deva and Doll{\'a}r, Piotr and Zitnick, C. Lawrence},
  booktitle={Proceedings of the European Conference on Computer Vision},
  year={2014}
}

@article{krishna2017visualgenome,
  title={Visual Genome: Connecting Language and Vision Using Crowdsourced Dense Image Annotations},
  author={Krishna, Ranjay and Zhu, Yuke and Groth, Oliver and Johnson, Justin and Hata, Kenji and Kravitz, Joshua and Chen, Stephanie and Kalantidis, Yannis and Li, Li-Jia and Shamma, David A. and Bernstein, Michael S. and Fei-Fei, Li},
  journal={International Journal of Computer Vision},
  year={2017}
}

@inproceedings{schuhmann2022laion5b,
  title={LAION-5B: An Open Large-Scale Dataset for Training Next Generation Image-Text Models},
  author={Schuhmann, Christoph and Beaumont, Romain and Vencu, Richard and Gordon, Cade and Wightman, Ross and Cherti, Mehdi and Coombes, Theo and Katta, Aarush and Mullis, Clayton and Wortsman, Mitchell and others},
  booktitle={Advances in Neural Information Processing Systems},
  year={2022}
}

@inproceedings{guss2019minerl,
  title={MineRL: A Large-Scale Dataset of Minecraft Demonstrations},
  author={Guss, William H. and Houghton, Brandon and Topin, Nicholay and Wang, Phillip and Codel, Cayden and Veloso, Manuela and Salakhutdinov, Ruslan},
  booktitle={Proceedings of the International Joint Conference on Artificial Intelligence},
  year={2019}
}

@inproceedings{fan2022minedojo,
  title={MineDojo: Building Open-Ended Embodied Agents With Internet-Scale Knowledge},
  author={Fan, Linxi and Wang, Guanzhi and Jiang, Yunfan and Mandlekar, Ajay and Yang, Yuncong and Zhu, Haoyi and Tang, Andrew and Huang, De-An and Zhu, Yuke and Anandkumar, Anima},
  booktitle={Advances in Neural Information Processing Systems},
  year={2022}
}

@article{he2025plaicraft,
  title={PLAICraft: Large-Scale Time-Aligned Vision-Speech-Action Dataset for Embodied AI},
  author={He, Yingchen and Weilbach, Christian D. and Wojciechowska, Martyna E. and Sun, Yiyuan and Zhang, Yuxuan and Wood, Frank},
  journal={arXiv preprint arXiv:2505.12707},
  year={2025}
}

@article{khazatsky2024droid,
  title={DROID: A Large-Scale In-The-Wild Robot Manipulation Dataset},
  author={Khazatsky, Alexander and Pertsch, Karl and Nair, Suraj and Balakrishna, Ashwin and Dasari, Sudeep and Karamcheti, Siddharth and Nasiriany, Soroush and Srirama, Mohan Kumar and Chen, Lawrence Yunliang and Ellis, Kirsty and others},
  journal={arXiv preprint arXiv:2403.12945},
  year={2024}
}

@article{padalkar2023openx,
  title={Open X-Embodiment: Robotic Learning Datasets and RT-X Models},
  author={Padalkar, Abhishek and Pooley, Alex and Jain, Ajay and Bewley, Alex and Herzog, Alexander and Irpan, Alex and Khazatsky, Alexander and Rai, Anant and Singh, Anikait and Brohan, Anthony and others},
  journal={arXiv preprint arXiv:2310.08864},
  year={2023}
}

@inproceedings{kirillov2023sam,
  title={Segment Anything},
  author={Kirillov, Alexander and Mintun, Eric and Ravi, Nikhila and Mao, Hanzi and Rolland, Chloe and Gustafson, Laura and Xiao, Tete and Whitehead, Spencer and Berg, Alexander C. and Lo, Wan-Yen and others},
  booktitle={Proceedings of the IEEE/CVF International Conference on Computer Vision},
  year={2023}
}

@inproceedings{caesar2020nuscenes,
  title={nuScenes: A Multimodal Dataset for Autonomous Driving},
  author={Caesar, Holger and Bankiti, Varun and Lang, Alex H. and Vora, Sourabh and Liong, Venice Erin and Xu, Qiang and Krishnan, Anush and Pan, Yu and Baldan, Giancarlo and Beijbom, Oscar},
  booktitle={Proceedings of the IEEE/CVF Conference on Computer Vision and Pattern Recognition},
  year={2020}
}

@inproceedings{sun2020waymo,
  title={Scalability in Perception for Autonomous Driving: Waymo Open Dataset},
  author={Sun, Pei and Kretzschmar, Henrik and Dotiwalla, Xerxes and Chouard, Aurelien and Patnaik, Vijaysai and Tsui, Paul and Guo, James and Zhou, Yin and Chai, Yuning and Caine, Benjamin and others},
  booktitle={Proceedings of the IEEE/CVF Conference on Computer Vision and Pattern Recognition},
  year={2020}
}

@inproceedings{yu2020bdd100k,
  title={BDD100K: A Diverse Driving Dataset for Heterogeneous Multitask Learning},
  author={Yu, Fisher and Chen, Haofeng and Wang, Xin and Xian, Wenqi and Chen, Yingying and Liu, Fangchen and Madhavan, Vashisht and Darrell, Trevor},
  booktitle={Proceedings of the IEEE/CVF Conference on Computer Vision and Pattern Recognition},
  year={2020}
}

@inproceedings{bruce2024genie,
title     = {Genie: Generative Interactive Environments},
author    = {Bruce, Jake and Dennis, Michael and Edwards, Ashley and Parker-Holder, Jack and Shi, Yuge and Hughes, Edward and Lai, Matthew and Mavalankar, Aditi and Steigerwald, Richie and Apps, Chris and Aytar, Yusuf and Bechtle, Sarah and Behbahani, Feryal and Chan, Stephanie and Heess, Nicolas and Gonzalez, Lucy and Osindero, Simon and Ozair, Sherjil and Reed, Scott and Zhang, Jingwei and Zolna, Konrad and Clune, Jeff and de Freitas, Nando and Singh, Satinder and Rockt{"a}schel, Tim},
booktitle = {Proceedings of the 41st International Conference on Machine Learning},
series    = {Proceedings of Machine Learning Research},
volume    = {235},
pages     = {4603--4623},
publisher = {PMLR},
year      = {2024}
}

@article{oasis2024,
title  = {Oasis: A Universe in a Transformer},
author = {{Decart} and Quevedo, Julian and McIntyre, Quinn and Campbell, Spruce and Chen, Xinlei and Wachen, Robert},
year   = {2024},
url    = {https://oasis-model.github.io/}
}

@inproceedings{valevski2024diffusion,
title     = {Diffusion Models Are Real-Time Game Engines},
author    = {Valevski, Dani and Leviathan, Yaniv and Arar, Moab and Fruchter, Shlomi},
booktitle = {International Conference on Learning Representations},
year      = {2025}
}

@inproceedings{alonso2024diffusion,
title     = {Diffusion for World Modeling: Visual Details Matter in Atari},
author    = {Alonso, Eloi and Jelley, Adam and Micheli, Vincent and Kanervisto, Anssi and Storkey, Amos and Pearce, Tim and Fleuret, Fran{\c{c}}ois},
booktitle = {Advances in Neural Information Processing Systems},
year      = {2024}
}

@inproceedings{yu2025gamefactory,
title     = {GameFactory: Creating New Games with Generative Interactive Videos},
author    = {Yu, Jiwen and Qin, Yiran and Wang, Xintao and Wan, Pengfei and Zhang, Di and Liu, Xihui},
booktitle = {Proceedings of the IEEE/CVF International Conference on Computer Vision},
pages     = {11590--11599},
month     = {October},
year      = {2025}
}

@article{guo2025mineworld,
title   = {MineWorld: A Real-Time and Open-Source Interactive World Model on Minecraft},
author  = {Guo, Junliang and Ye, Yang and He, Tianyu and Wu, Haoyu and Jiang, Yushu and Pearce, Tim and Bian, Jiang},
journal = {arXiv preprint arXiv:2504.08388},
year    = {2025}
}

@article{tang2025hunyuangamecraft,
title   = {Hunyuan-GameCraft: High-Dynamic Interactive Game Video Generation with Hybrid History Condition},
author  = {Li, Jiaqi and Tang, Junshu and Xu, Zhiyong and Wu, Longhuang and Zhou, Yuan and Shao, Shuai and Yu, Tianbao and Cao, Zhiguo and Lu, Qinglin},
journal = {arXiv preprint arXiv:2506.17201},
year    = {2025}
}

@article{tang2025hunyuangamecraft2,
title   = {Hunyuan-GameCraft-2: Instruction-Following Interactive Game World Model},
author  = {Tang, Junshu and Liu, Jiacheng and Li, Jiaqi and Wu, Longhuang and Yang, Haoyu and Zhao, Penghao and Gong, Siruis and Yuan, Xiang and Shao, Shuai and Lu, Qinglin},
journal = {arXiv preprint arXiv:2511.23429},
year    = {2025}
}

@article{mao2025yume,
title   = {Yume: An Interactive World Generation Model},
author  = {Mao, Xiaofeng and Lin, Shaoheng and Li, Zhen and Li, Chuanhao and Peng, Wenshuo and He, Tong and Pang, Jiangmiao and Chi, Mingmin and Qiao, Yu and Zhang, Kaipeng},
journal = {arXiv preprint arXiv:2507.17744},
year    = {2025}
}

@inproceedings{mao2025yume15,
title     = {Yume1.5: A Text-Controlled Interactive World Generation Model},
author    = {Mao, Xiaofeng and Li, Zhen and Li, Chuanhao and Xu, Xiaojie and Ying, Kaining and Zhang, Kaipeng},
booktitle = {Proceedings of the IEEE/CVF Conference on Computer Vision and Pattern Recognition},
pages     = {7752--7761},
month     = {June},
year      = {2026}
}

@inproceedings{feng2024matrix,
title     = {The Matrix: Infinite-Horizon World Generation with Real-Time Moving Control},
author    = {Feng, Ruili and Zhang, Han and Shu, Zhilei and Yang, Zhantao and Tang, Longxiang and Wang, Zhicai and Zheng, Andy and Xiao, Jie and Liu, Zhiheng and Chu, Ruihang and Huang, Yukun and Liu, Yu and Zhang, Hongyang},
booktitle = {Advances in Neural Information Processing Systems},
year      = {2025}
}

@article{zhang2025matrixgame,
title   = {Matrix-Game: Interactive World Foundation Model},
author  = {Zhang, Yifan and Peng, Chunli and Wang, Boyang and Wang, Puyi and Zhu, Qingcheng and Kang, Fei and Jiang, Biao and Gao, Zedong and Li, Eric and Liu, Yang and Zhou, Yahui},
journal = {arXiv preprint arXiv:2506.18701},
year    = {2025}
}

@article{wang2026matrixgame3,
title   = {Matrix-Game 3.0: Real-Time and Streaming Interactive World Model with Long-Horizon Memory},
author  = {Wang, Zile and Liu, Zexiang and Li, Jaixing and Huang, Kaichen and Xu, Baixin and Kang, Fei and An, Mengyin and Wang, Peiyu and Jiang, Biao and Wei, Yichen and Xietian, Yidan and Pei, Jiangbo and Hu, Liang and Jiang, Boyi and Xue, Hua and Wang, Zidong and Sun, Haofeng and Li, Wei and Ouyang, Wanli and He, Xianglong and Liu, Yang and Li, Yangguang and Zhou, Yahui},
journal = {arXiv preprint arXiv:2604.08995},
year    = {2026}
}

@inproceedings{xiao2025worldmem,
title     = {WorldMem: Long-Term Consistent World Simulation with Memory},
author    = {Xiao, Zeqi and Lan, Yushi and Zhou, Yifan and Ouyang, Wenqi and Yang, Shuai and Zeng, Yanhong and Pan, Xingang},
booktitle = {Advances in Neural Information Processing Systems},
year      = {2025}
}

@article{hong2025relic,
title   = {RELIC: Interactive Video World Model with Long-Horizon Memory},
author  = {Hong, Yicong and Mei, Yiqun and Ge, Chongjian and Xu, Yiran and Zhou, Yang and Bi, Sai and Hold-Geoffroy, Yannick and Roberts, Mike and Fisher, Matthew and Shechtman, Eli and Sunkavalli, Kalyan and Liu, Feng and Li, Zhengqi and Tan, Hao},
journal = {arXiv preprint arXiv:2512.04040},
year    = {2025}
}

@article{shang2025longscape,
title   = {LongScape: Advancing Long-Horizon Embodied World Models with Context-Aware MoE},
author  = {Shang, Yu and Jin, Lei and Ma, Yiding and Zhang, Xin and Gao, Chen and Wu, Wei and Li, Yong},
journal = {arXiv preprint arXiv:2509.21790},
year    = {2025}
}

@article{nam2026worldcam,
title   = {WorldCam: Interactive Autoregressive 3D Gaming Worlds with Camera Pose as a Unifying Geometric Representation},
author  = {Nam, Jisu and Hong, Yicong and Huang, Chun-Hao Paul and Liu, Feng and Lee, JoungBin and Kim, Jiyoung and Jin, Siyoon and Lee, Yunsung and Jung, Jaeyoon and Choi, Suhwan and Kim, Seungryong and Zhou, Yang},
journal = {arXiv preprint arXiv:2603.16871},
year    = {2026}
}

@misc{worldscape2026,
title        = {WorldScape: An Efficient Real-Time Interactive World Model with Universal Control},
author       = {{WorldScape Team}},
year         = {2026},
howpublished = {Technical report},
url          = {https://worldscape.io/}
}

@misc{brooks2024sora,
title = {Video Generation Models as World Simulators},
author = {Brooks, Tim and Peebles, Bill and Holmes, Connor and DePue, Will and Guo, Yufei and Jing, Li and Schnurr, David},
year = {2024},
howpublished = {OpenAI technical report},
url = {https://openai.com/index/video-generation-models-as-world-simulators/}
}

@article{polyak2024moviegen,
title = {Movie Gen: A Cast of Media Foundation Models},
author = {Polyak, Adam and Zohar, Amit and Brown, Andrew and Tjandra, Andros and Sinha, Animesh and Lee, Ann and Vyas, Apoorv and Shi, Bowen and Ma, Chih-Yao and Chuang, Ching-Yao and others},
journal = {arXiv preprint arXiv:2410.13720},
year = {2024}
}

@article{valevski2024gamengen,
title = {Diffusion Models Are Real-Time Game Engines},
author = {Valevski, Dani and Leviathan, Yaniv and Arar, Moab and Fruchter, Shlomi},
journal = {arXiv preprint arXiv:2408.14837},
year = {2024}
}

@inproceedings{kim2024openvla,
title = {{OpenVLA}: An Open-Source Vision-Language-Action Model},
author = {Kim, Moo Jin and Pertsch, Karl and Karamcheti, Siddharth and Xiao, Ted and Balakrishna, Ashwin and Nair, Suraj and Rafailov, Rafael and Foster, Ethan P. and Sanketi, Pannag R. and Vuong, Quan and Kollar, Thomas and Burchfiel, Benjamin and Tedrake, Russ and Sadigh, Dorsa and Levine, Sergey and Liang, Percy and Finn, Chelsea},
booktitle = {Proceedings of The 8th Conference on Robot Learning},
pages = {2679--2713},
year = {2025},
volume = {270},
series = {Proceedings of Machine Learning Research},
publisher = {PMLR}
}

@article{black2024pi0,
title = {$\pi_0$: A Vision-Language-Action Flow Model for General Robot Control},
author = {Black, Kevin and Brown, Noah and Driess, Danny and Esmail, Adnan and Equi, Michael and Finn, Chelsea and Fusai, Niccolo and Groom, Lachy and Hausman, Karol and Ichter, Brian and Jakubczak, Szymon and Jones, Tim and Ke, Liyiming and Levine, Sergey and Li-Bell, Adrian and Mothukuri, Mohith and Nair, Suraj and Pertsch, Karl and Shi, Lucy Xiaoyang and Tanner, James and Vuong, Quan and Walling, Anna and Wang, Haohuan and Zhilinsky, Ury},
journal = {arXiv preprint arXiv:2410.24164},
year = {2024}
}

@article{huang2025voyager,
  title={Voyager: Long-range and world-consistent video diffusion for explorable 3d scene generation},
  author={Huang, Tianyu and Zheng, Wangguandong and Wang, Tengfei and Liu, Yuhao and Wang, Zhenwei and Wu, Junta and Jiang, Jie and Li, Hui and Lau, Rynson and Zuo, Wangmeng and others},
  journal={ACM Transactions on Graphics (TOG)},
  volume={44},
  number={6},
  pages={1--15},
  year={2025},
  publisher={ACM New York, NY, USA}
}

@article{team2025hunyuanworld,
  title={{HunyuanWorld} 1.0: Generating Immersive, Explorable, and Interactive {3D} Worlds from Words or Pixels},
  author={{HunyuanWorld Team} and Wang, Zhenwei and Liu, Yuhao and Wu, Junta and Gu, Zixiao and Wang, Haoyuan and Zuo, Xuhui and Huang, Tianyu and Li, Wenhuan and Zhang, Sheng and others},
  journal={arXiv preprint arXiv:2507.21809},
  year={2025}
}

\end{document}